\documentclass[runningheads]{llncs}
\usepackage[T1]{fontenc}
\usepackage{graphicx}
\usepackage{makecell}
\usepackage{multirow}
\usepackage{amsmath}
\usepackage{amsfonts}
\usepackage{mathrsfs}
\usepackage[bookmarks=true]{hyperref}
\usepackage{lipsum}
\usepackage{xcolor}
\usepackage{bbding}
\usepackage{hyperref}

\definecolor{gray}{RGB}{128,128,128}
\SetLipsumParListSurrounders{\begingroup\color{gray}}{\endgroup}

\definecolor{myred}{RGB}{255,0,0}
\definecolor{mygreen}{RGB}{0,176,80}

\usepackage{xspace}
\makeatletter
\DeclareRobustCommand\onedot{\futurelet\@let@token\@onedot}
\def\@onedot{\ifx\@let@token.\else.\null\fi\xspace}

\def\ie{\emph{i.e}\onedot}

\makeatother

\usepackage{color}

\begin{document}
\title{Towards Clinician-Preferred Segmentation: Leveraging Human-in-the-Loop for Test Time Adaptation in Medical Image Segmentation}
\titlerunning{Leveraging Human-in-the-Loop for Test Time Adaptation}
%
\author{Shishuai Hu\inst{1} \and
Zehui Liao\inst{1} \and
Zeyou Liu\inst{1} \and
Yong Xia\inst{1,2,3}\Envelope}
\authorrunning{S. Hu et al.}
%
\institute{
National Engineering Laboratory for Integrated Aero-Space-Ground-Ocean Big Data Application Technology, School of Computer Science and Engineering, Northwestern Polytechnical University, Xi’an 710072, China \\
\email{yxia@nwpu.edu.cn}
\and
Ningbo Institute of Northwestern Polytechnical University, Ningbo 315048, China
\and
Research and Development Institute of Northwestern Polytechnical University in Shenzhen, Shenzhen 518057, China
}
\maketitle              
%

\begin{abstract}
Deep learning-based medical image segmentation models often face performance degradation when deployed across various medical centers, largely due to the discrepancies in data distribution. 
Test Time Adaptation (TTA) methods, which adapt pre-trained models to test data, have been employed to mitigate such discrepancies.
However, existing TTA methods primarily focus on manipulating Batch Normalization (BN) layers or employing prompt and adversarial learning, which may not effectively rectify the inconsistencies arising from divergent data distributions.
In this paper, we propose a novel Human-in-the-loop TTA (HiTTA) framework that stands out in two significant ways. First, it capitalizes on the largely overlooked potential of clinician-corrected predictions, integrating these corrections into the TTA process to steer the model towards predictions that coincide more closely with clinical annotation preferences. Second, our framework conceives a divergence loss, designed specifically to diminish the prediction divergence instigated by domain disparities, through the careful calibration of BN parameters.
Our HiTTA is distinguished by its dual-faceted capability to acclimatize to the distribution of test data whilst ensuring the model's predictions align with clinical expectations, thereby enhancing its relevance in a medical context. 
Extensive experiments on a public dataset underscore the superiority of our HiTTA over existing TTA methods, emphasizing the advantages of integrating human feedback and our divergence loss in enhancing the model's performance and adaptability across diverse medical centers.

\end{abstract}
\section{Introduction}
Deploying a well-trained deep learning-based medical image segmentation model to a new medical center often faces performance degradation due to the discrepancies in data distribution across different medical centers~\cite{LITJENS201760,IMGCHAR,guan2021domain,xie_survey_2021}. 
Test Time Adaptation (TTA)~\cite{schneider2020improving,wang2020tent,wang2022continual,lim2023ttn,niu2023towards,zhang2023domainadaptor,lee2024entropy,chen2023each} remedies this issue by fine-tuning the pre-trained model using current test data, thus adapting the model to the test data distribution during the inference stage.

Most TTA methods focus on utilizing~\cite{hu2022prosfda,chen2023each,yang2022source} or adjusting~\cite{schneider2020improving,bronskill2020tasknorm,wang2020tent,lim2023ttn,niu2023towards,lee2024entropy} the Batch Normalization (BN) layers in the segmentation models to better match the test data distribution.
This is largely due to the fact that the statistics and learnable parameters housed in the BN layers mirror the overall data distribution.
With these statistics stored within the BN layers of the pre-trained model, techniques like prompt learning~\cite{hu2022prosfda,chen2023each} and adversarial training~\cite{yang2022source} are employed to align the current test data with the source data distribution.
Despite the statistical alignment improvements, these methods often suffer from unreliable data adjustments, since such modifications can jeopardize segmentation tasks.
Another approach involves re-estimate BN statistics using current test data~\cite{schneider2020improving,bronskill2020tasknorm}.
The updated statistics can then be used for forward propagation for prediction.
Although avoiding the modification of test data, these methods encounter issues related to inaccurate estimation due to insufficient test data for estimation.

To further enhance the model's fit with the test data distribution, researchers have recently turned their focus to fine-tuning learnable parameters in BN layers, in addition to re-estimating statistics~\cite{wang2020tent,lim2023ttn,niu2023towards,lee2024entropy}.
These methods employ the Shannon entropy~\cite{shannon1948mathematical,wang2020tent,lim2023ttn} or its variants~\cite{niu2023towards,lee2024entropy} of the model prediction as the optimization objective, presuming that the high entropy is mainly caused by the discrepancies in data distribution.
While these methods have furthered the adaptability of learnable BN parameters, the reliability of the entropy used can often suffer, particularly with medical image segmentation where irregular shapes or ambiguous borders of segmentation targets can also cause high entropy~\cite{wang2019boundary}.
To address this, we propose a new divergence loss that is less susceptible to segmentation targets.
Specifically, given a test sample, the model generates predictions for both the test data and its style-augmented counterparts~\cite{hu2023devil}.
These augmented samples have the same structure as the test data but different styles.
The divergence loss is defined as the divergence between the predictions.
This measure provides a more accurate assessment of how domain discrepancies influence model performance, and minimizing it enables better adaptation of the model.

\begin{figure}[t]
    \centering
    \includegraphics[width=1\textwidth]{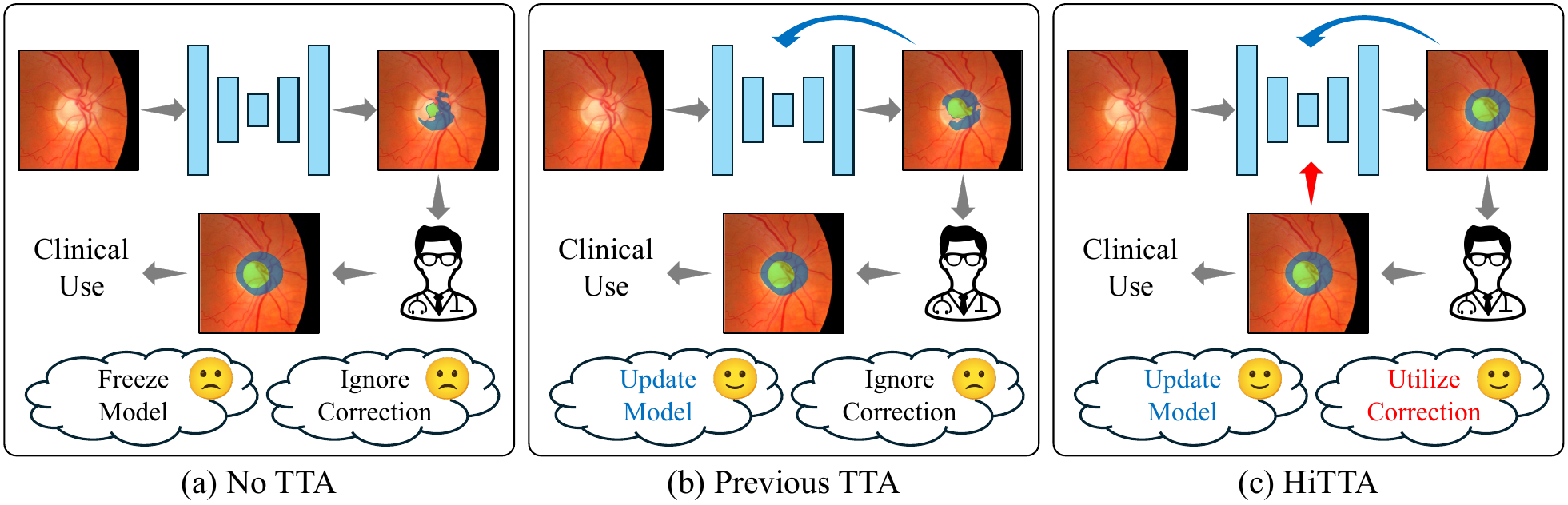}
    \caption{Comparison of (a) No TTA, (b) Previous TTA Methods, and (c) Proposed HiTTA. Gray arrows show data flow, blue arrows indicate model updates using predictions, and red arrow highlights optimization with clinician corrections. Unlike previous methods, HiTTA incorporates human-in-the-loop feedback for enhanced performance.
    }
    \label{fig:motivation}
  \end{figure}

Although the performance degradation caused by the data distribution discrepancies can be mitigated, predictions from the adapted model still require modifications from clinicians for clinical suitability,
since the annotation preference of the model's training data can significantly differ from that required by clinicians~\cite{liao2024modeling,liao2023transformer,suetens2017fundamentals}.
In practice, clinicians usually modify model predictions until they are satisfactory, and then use these corrected masks for further clinical tasks (see Fig.~\ref{fig:motivation} (a)).
These corrected predictions, \ie, human feedback, however, are rarely~\cite{wang2020tent,lim2023ttn,niu2023towards,lee2024entropy,chen2023each} or only partially~\cite{gui2024active} utilized in TTA workflow (see Fig.~\ref{fig:motivation} (b)).
To align with the annotation preference of the clinician and continue improving the model performance, we suggest utilizing human feedback to guide model optimization (Fig.~\ref{fig:motivation} (c)).

In this paper, we propose a new optimization objective for TTA and incorporate it into a novel \textbf{H}uman-\textbf{i}n-the-loop \textbf{T}est \textbf{T}ime \textbf{A}daptation (\textbf{HiTTA}) framework for medical image segmentation.
In HiTTA, TTA is divided into three stages: pre-inference, inference, and post-inference.
The pre-inference stage involves augmenting the test sample to style-augmented counterparts~\cite{hu2023devil} and using them to generate varied predictions.
We then use the proposed divergence loss to optimize the model's BN parameters, making the model less susceptible to the discrepancies between the test data and unseen training data.
The inference stage involves feeding the test data into the optimized model to generate a prediction which can then be corrected by a clinician. In the post-inference stage, we utilize these corrections for supervising model optimization. This enables the model to better mimic the clinician's annotation preference.
The proposed HiTTA has been evaluated against other TTA methods on a public dataset and improved performance has been achieved.

To summarize, our contributions are three-fold.
\begin{itemize}
    \item We introduce a divergence loss to quantify the impact of domain discrepancies on model performance, enabling more effective fine-tuning of BN parameters and improved adaptation to test data.
    \item We propose to incorporate clinician corrections into the TTA framework to guide model optimization, aligning model prediction with clinical annotation preference and thus improving the model's clinical utility.
    \item We conduct extensive experiments on a cross-domain, multi-annotator, and joint optic disc (OD) / optic cup (OC) segmentation dataset to demonstrate the effectiveness of our proposed method.
\end{itemize}

\section{Method}
\begin{figure}[t]
    \centering
    \includegraphics[width=1\textwidth]{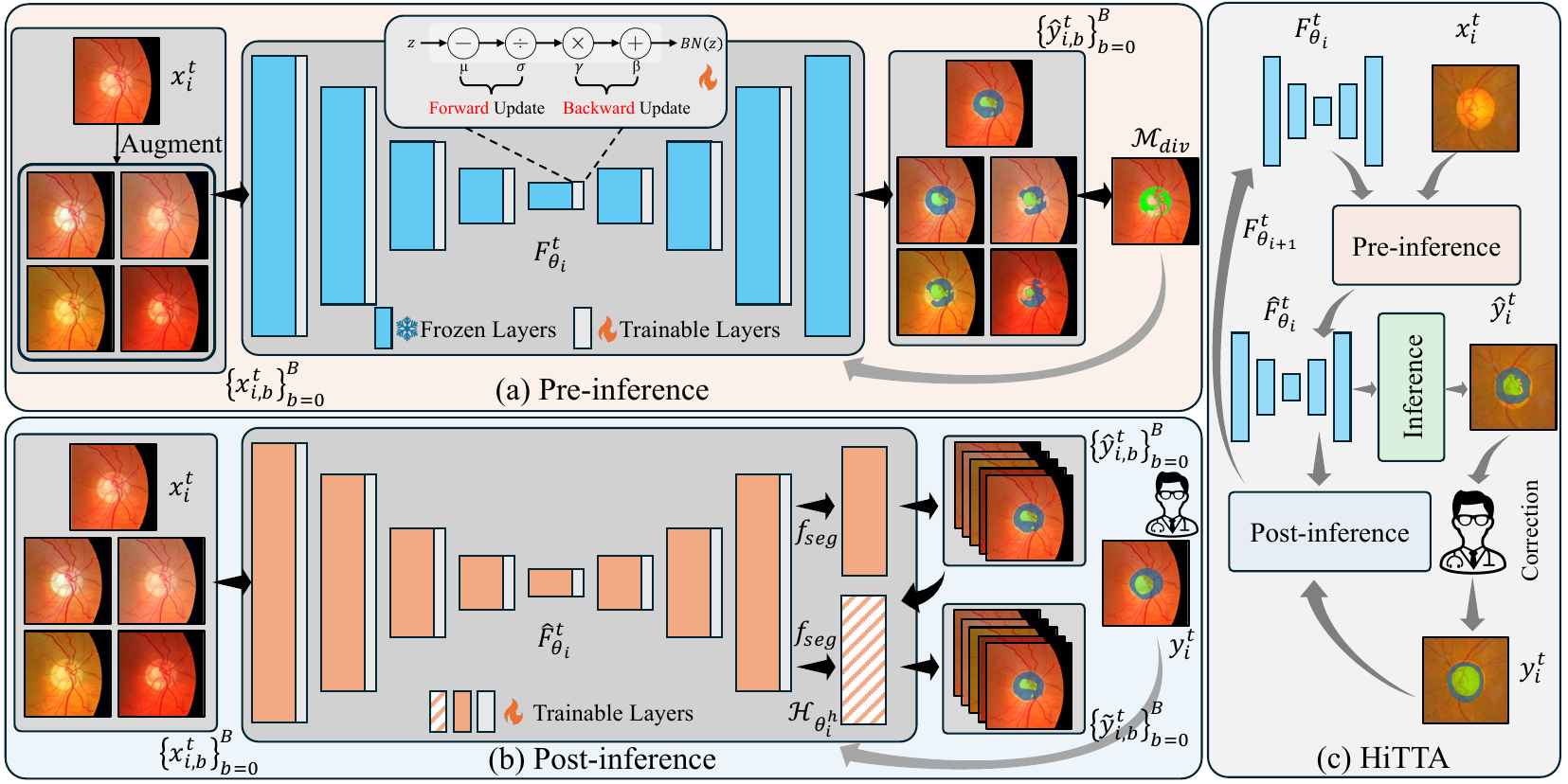}
    \caption{Illustration of proposed HiTTA framework. It is mainly composed of (a) Pre-inference Stage, inference stage, and (b) Post-inference Stage. (c) shows the workflow of the proposed HiTTA framework.}
    \label{fig:overview}
  \end{figure}

\subsection{Problem Definition and Method Overview}
Let a segmentation model trained on source domain data $\mathbb{D}^s = \{(x_n^s, y_n^s)_{n=1}^N\}$ be denoted as $F_{\theta}^s: x \to y$.
During inference, the test model $F_{\theta_1}^t$ is initialized with $F_{\theta}^s$.
Given a test sample $x_i^t$ from the target domain data $\mathbb{D}^t=\{x^t\}$, the model generates a prediction $\hat{y}_i^t = F_{\theta_i}^t(x_i^t)$, which could be distorted by domain discrepancies.
The first goal is to adapt $F_{\theta_i}^t$ to $\hat{F}_{\theta_i}^t$, using $x_i^t$ such that the adapted model $\hat{F}_{\theta_i}^t$ can generate an improved $\hat{y}_i^t$ that will be presented to a clinician.
The corrections from this prediction $\hat{y}_i^t$ to $y_i^t$ by the clinician, also equips us to further adapt $\hat{F}_{\theta_i}^t$ to $F_{\theta_{i+1}}^t$ using $y_i^t$, which is the second goal.
Consequently, the model not only performs better on the ensuing test data $x_{i+1}^t$, but also effectively mimics the clinical annotation preference.

To achieve these goals, we therefore align data distribution in the pre-inference stage and model annotation preference in the post-inference stage.
During the pre-inference stage, the test sample $x_i^t$ is augmented to obtain its augmented counterparts $\{x_{i, b}^t\}_{b=1}^B$ via style-augmentation~\cite{hu2023devil}.
Feeding the test input batch $\{x_{i, b}^t\}_{b=0}^B$, where $x_{i, 0}^t = x_i^t$, into the model produces predictions $\{\hat{y}_{i, b}^t\}_{b=0}^B$.
The divergence loss $\mathcal{L}_{div}$ is then calculated using Equation~\ref{eq:div}, minimized to update the BN parameters in $F_{\theta_i}^t$, resulting in the adapted model $\hat{F}_{\theta_i}^t$, which generates $\hat{y}_i^t$. This is then corrected by clinician to $y_i^t$.
During the post-inference stage, an additional preference-involved segmentation head $\mathcal{H}_{\theta_i^h}$ is integrated into the model to mimic the annotation preference observed in $y_i^t$, hence updating $\mathcal{H}_{\theta_i^h}$ and $\hat{F}_{\theta_i}^t$ to yield $F_{\theta_{i+1}}^t$.
The diagram of HiTTA is shown in Fig.~\ref{fig:overview}. We now delve into its details.

\subsection{Pre-inference Stage}
Let an input of a BN layer be $z \in \mathbb{R}^{B\times C \times H \times W}$, the output of the BN layer can be formulated as:
\begin{equation}
    \text{BN}(z) = \gamma \frac{z - \mu}{\sqrt{\sigma^2 + \epsilon}} + \beta,
\end{equation}
where $\mu$ and $\sigma^2$ are the mean and variance of $z$ in the mini-batch, $\gamma$ and $\beta$ are learnable scale and bias factors, and $\epsilon$ is a small constant to avoid division by zero.

During the training of source model, $\mu^s$ and $\sigma^s$ are the moving average of the mini-batch statistics, and $\gamma^s$ and $\beta^s$ are learned via back-propagation.
Under an i.i.d assumption, $\mu^s$ and $\sigma^s$ can be regarded as population statistics, with $\gamma^s$ and $\beta^s$ serving as the optimal scale and bias factors for both training and test data.
In TTA setting, where the i.i.d assumption is less likely to be held, directly applying these BN parameters can degrade the test data performance.
Therefore, we adjust the estimated $\mu^s$ and $\sigma^s$ with the mini-test-batch statistics $\mu^t$ and $\sigma^t$.
Since only one test sample can be accessed at a time, the estimated $\mu^t$ and $\sigma^t$ may not accurately reflect the test data distribution, making model adaptation challenging.
To alleviate this, we augment the test sample $x_i^t$ using style-augmentation techniques~\cite{hu2023devil}, including gaussian noise addition, gaussian blur, brightness adjusting, contrast adjusting, and gamma transform.
Then we stack the test sample with these counterparts to form a test batch $\{x_{i, b}^t\}_{b=0}^B$ for re-estimation of $\mu^t$ and $\sigma^t$.

With the lack of ground truth, it is not possible to directly optimize $\gamma^t$ and $\beta^t$.
We introduce the divergence loss $\mathcal{L}_{div}$ to quantify the impact of domain discrepancies on model performance.
Given the augmented test batch $\{x_{i, b}^t\}_{b=0}^B$, the respective predictions $\{\hat{y}_{i, b}^t\}_{b=0}^B$ are obtained using $F_{\theta_i}^t$.
The prediction divergence map is then calculated as:
\begin{equation}
    \mathcal{M}_{div} = \frac{1}{B}\sum_{b=0}^B \|\hat{y}_{i, b}^t - \frac{1}{B}\sum_{b=0}^B \hat{y}_{i, b}^t\|_2,
\label{eq:div}
\end{equation}
where $\mathcal{L}_{div}$ is designed as the average of $\mathcal{M}_{div}$.
The adapted model $\hat{F}_{\theta_i}^t$ is obtained by minimizing $\mathcal{L}_{div}$ to update $\gamma^t$ and $\beta^t$.

\subsection{Post-inference Stage}
Given the corrected segmentation mask $y_i^t$, an additional segmentation head $\mathcal{H}_{\theta_i^h}$, which consists of two stacked convolutional layers with $3 \times 3$ kernels, is appended to the model to mimic the annotation preference within it.
The input to $\mathcal{H}_{\theta_i^h}$ is the concatenation of the segmentation feature map $f_{seg}$ before the main segmentation head and $\hat{y}_i^t$ (see Fig.~\ref{fig:overview} (b)).
The output of $\mathcal{H}_{\theta_i^h}$ can be represented as:
\begin{equation}
    \widetilde{y}_i^t = \mathcal{H}_{\theta_{i}^h}(concat(f_{seg}, \hat{y}_i^t)).
\end{equation}
Both $\{\theta_{i}^h, \theta_{i}\}$ can be optimized using the combination of the Dice loss and the cross entropy loss:
\begin{equation}
    \mathcal{L}_{seg} = \mathcal{L}_{dice}(\widetilde{y}_i^t, y_i^t) + \mathcal{L}_{ce}(\widetilde{y}_i^t, y_i^t) + \mathcal{L}_{dice}(\hat{y}_i^t, y_i^t) + \mathcal{L}_{ce}(\hat{y}_i^t, y_i^t).
\end{equation}
The test sample $x_i^t$ is again augmented to $\{x_{i, b}^t\}_{b=0}^B$, and $\mathcal{M}_{div}$ is calculated.
To compel the model towards more consistent predictions, we use $1 + \mathcal{M}_{div}$ as a weighted map to amplify $\mathcal{L}_{seg}$.
To circumvent catastrophic forgetting, we set the learning rate of $\theta_{i}$ to be smaller than that of $\theta_{i}^h$.

\section{Experiments and Results}
\subsection{Materials and Evaluation Metric}
The RIGA+ dataset~\cite{hu2022domain,almazroa2018retinal,decenciere2014feedback} is a multi-domain joint OC/OD segmentation dataset annotated by six ophthalmologists. The dataset includes data from five medical centers, namely BinRushed, BASE1, BASE2, BASE3, and Magrabia.
The respective domains contain 195, 173, 148, 133, and 95 labeled images. 
To simulate differing annotation preference, we assigned the annotations of the first five ophthalmologists (R1-R5) to each domain, respectively, employed BinRushed as the source domain, and tested on other four domains sequentially.
The Dice Similarity Coefficient (DSC) served as our primary metric to measure segmentation performance.

\subsection{Implementation Details}
The images were normalized by subtracting the mean and dividing by the standard deviation.
We adopted nnUNet~\cite{isensee2021nnu} as the backbone.
\textbf{Training source model:} The source model was trained over 100 epochs.
The SGD optimizer with an inaugural learning rate $lr_0$ of 0.01 and a momentum of 0.99 was used.
The learning rate was decayed per the polynomial rule $lr = lr_0 \times (1 - \frac{e}{E})^{0.9}$, where $e$ is the current epoch and $E$ is the maximum epoch.
We set the batch size to 12 while fixing an input size of $512\times 512$.
\textbf{Test time adaptation:} Test time adaptation was executed over 20 iterations. 
The input size was same as the source model training.
The augmented batch size $B$ is set to 6 for OD/OC segmentation.
The learning rate is set to 0.01 to optimize the $\gamma^t$ and $\beta^t$, 0.01 to optimize the $\theta_{i}^h$, and 0.001 to optimize $\theta_{i}$.
We presented both segmentation head predictions to clinicians during inference due to unreliable initial predictions from the preference-involved segmentation head. The better prediction was selected for evaluation.

\begin{table}[t]
\setlength\tabcolsep{1pt}
\centering
\caption{
Performance (DSC\%) of HiTTA and eight existing methods on the joint OD/OC segmentation dataset. The average DSC of OD and OC are reported. R1-R5 denote the five annotation preferences. R* denotes the average performance over the domain-specific annotation preferences (\ie, R2-R5). The best performance among the proposed HiTTA and six competing methods is highlighted with \textbf{bold}.
}
\label{tab:od/oc}
    \begin{tabular}{l|c|c|c|c|c|c|c|c|c|c}
        \hline \hline
    \multirow{2}{*}{Methods}  & \multicolumn{2}{c|}{BASE1} & \multicolumn{2}{c|}{BASE2} & \multicolumn{2}{c|}{BASE3} & \multicolumn{2}{c|}{Magrabia} & \multicolumn{2}{c}{Average} \\ \cline{2-11}
                              & vs. R1      & vs. R2      & vs. R1      & vs. R3      & vs. R1      & vs. R4      & vs. R1        & vs. R5       & vs. R1  & vs. R*  \\ \hline
    \hline
    Intra-Domain              & 89.22 & 89.51 & 90.09 & 92.29 & 86.90 & 91.44 & 86.69 & 87.33 & 88.22 & 90.14     \\ \hline
    w/o TTA                   & 88.16 & 86.55 & 75.23 & 72.51 & 87.83 & 85.05 & 81.54 & 79.54 & 83.19 & 80.91     \\ \hline
    \hline
    TBN~\cite{schneider2020improving}        & 87.98 & 86.11 & 86.95 & 84.27 & 89.33 & 84.77 & 85.19 & 83.20 & 87.36 & 84.59     \\ \hline
    TENT~\cite{wang2020tent}          & 88.14 & 86.30 & 87.93 & 84.98 & 89.57 & 86.47 & 85.46 & 83.26 & 87.78 & 85.25     \\ \hline
    TTN~\cite{lim2023ttn}           & 89.27 & 87.57 & 87.71 & 83.96 & 88.86 & 88.64 & 86.99 & 84.67 & 88.21 & 86.21     \\ \hline
    Do. Ad.~\cite{zhang2023domainadaptor} & 89.21 & 88.65 & 85.40 & 82.71 & \textbf{89.89} & 86.19 & 87.96 & 84.03 & 88.12 & 85.40     \\ \hline
    DeYO~\cite{lee2024entropy}          & 89.11 & 88.31 & 88.71 & 84.70 & 89.72 & 88.93 & 87.49 & 85.28 & 88.76 & 86.81     \\ \hline
    VPTTA~\cite{chen2023each}         & 89.28 & 87.65 & 87.87 & 84.13 & 88.87 & 88.64 & 87.30 & 84.95 & 88.33 & 86.34     \\ \hline
    Ours (HiTTA)       & \textbf{89.61} & \textbf{90.71} & \textbf{90.18} & \textbf{93.30} & 88.53 & \textbf{91.11} & \textbf{89.77} & \textbf{89.44} & \textbf{89.52} & \textbf{91.14}    \\ \hline
    \hline
    \end{tabular}
\end{table}

\begin{figure}[t]
    \centering
    \includegraphics[width=1\textwidth]{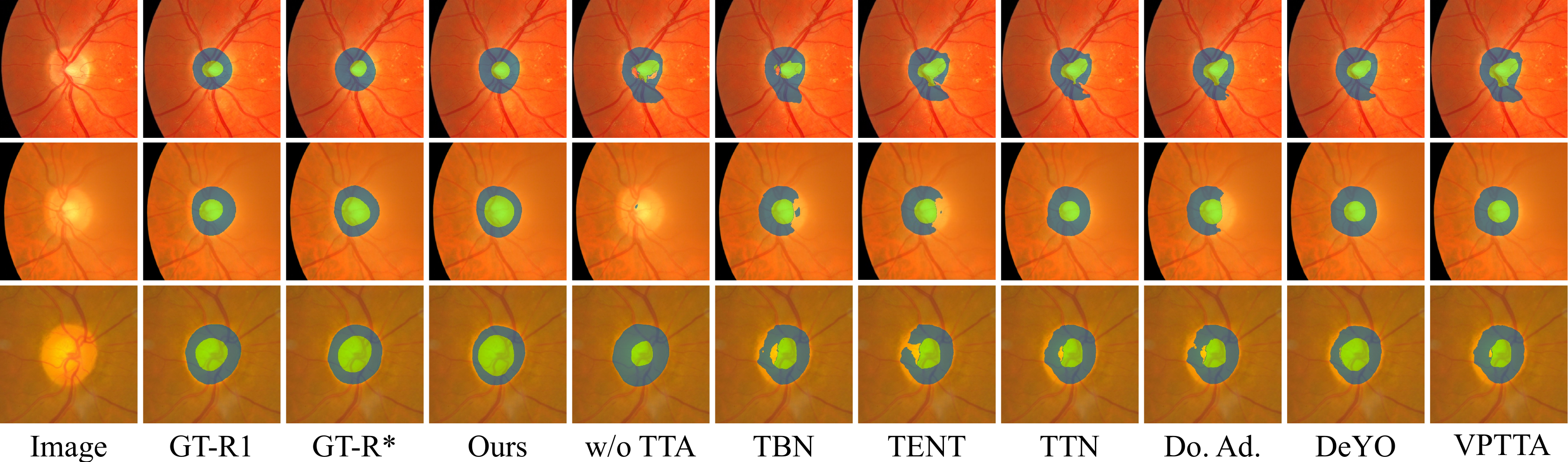}
    \caption{Visualization of segmentation masks predicted by HiTTA and seven existing methods, together with ground truth (GT-R1 and GT-R*).}
    \label{fig:compare}
  \end{figure}

\subsection{Comparative Experiments}
The proposed HiTTA was compared to six popular TTA methods, \ie, TBN (NeurIPS 2020)~\cite{schneider2020improving}, TENT (ICLR 2020)~\cite{wang2020tent}, TTN (ICLR 2023)~\cite{lim2023ttn}, DomainAdaptor (ICCV 2023)~\cite{zhang2023domainadaptor}, DeYO (ICLR 2024)~\cite{lee2024entropy}, and VPTTA (CVPR 2024)~\cite{chen2023each}.
We re-implemented all the competing methods under the same TTA setting, and reported their best performance.
Note that previous TTA methods were all principally designed for data distribution adaptation, hence we also reported their performance when equipped with human feedback in the second block of Table~\ref{tab:ablation}.
The results, displayed in Table~\ref{tab:od/oc}, reveal that HiTTA outperforms all competing methods, whether evaluated using the source domain annotation preference or target domain-specific annotation preference.
By contrast, previous TTA methods could only enhance performance when evaluated using source domain annotation preference.
This emphasizes the benefits of incorporating human feedback into the TTA workflow.
We visualized the segmentation predictions generated by HiTTA and other methods in Fig.~\ref{fig:compare}. 
It shows that the segmentation masks generated by HiTTA most closely align with GT-R*, confirming that HiTTA can generate more human-like predictions.

\subsection{Ablation Analysis}
The primary contributors of HiTTA were the divergence loss in the pre-inference stage and the newly incorporated post-inference stage.
To investigate their effectiveness, we conducted an ablation analysis, removing each contributor and reporting the performance of the variants.
The results, in the first block of Table~\ref{tab:ablation}, demonstrate a lower performance than that of HiTTA, affirming the effectiveness of the proposed divergence loss and post-inference stage.
Especially, the significance of human feedback in the TTA process becomes evident in the performance drop after its removal in the post-inference stage.

\noindent
\textbf{Analysis of Divergence Loss.}
We compared the updating of BN statistics and learnable parameters via minimizing proposed divergence loss or prediction entropy (TENT~\cite{wang2020tent}), with results listed in Table~\ref{tab:ablation} (`Ours w/o HF') and Table~\ref{tab:od/oc} (TENT). Comparatively, the proposed divergence loss proved more effective.

\noindent
\textbf{Analysis of Human Feedback.} 
Our modulary designed post-inference stage exhibits compatibility with other Test Time Adaptation (TTA) methods, allowing for easy integration.
We incorporated human feedback into TENT~\cite{wang2020tent} and DeYO~\cite{lee2024entropy} and reported the performance in the second block of Table~\ref{tab:ablation}.
Both methods showed enhanced performance after equipping with human feedback. This enhancement was particularly noticeable when compared with domain-specific annotation preferences, underlining the efficacy of the proposed post-inference stage.

To identify the optimal design of the preference-involved segmentation head, we executed additional experiments and reported the results in the third block of Table~\ref{tab:ablation}.
It shows that superior performance was observed with the use of two layers of $3\times 3$ kernels.
In the post-inference stage, we adopted $\mathcal{M}_{div}$ as the weighted map to boost $\mathcal{L}_{seg}$.
When contrasted with the entropy map utilized as the weighted map, the use of $\mathcal{M}_{div}$ procured superior performance, as stipulated in the last block of Table~\ref{tab:ablation}.

\begin{table}[t]
    \setlength\tabcolsep{0.5pt}
    \centering
    \caption{
    Performance (DSC\%) of proposed HiTTA and its variants on the joint OD/OC segmentation dataset. The average DSC of OD and OC are reported. R1-R5 denote the five annotation preferences. R* denotes the average performance over the domain-specific annotation preferences (\ie, R2-R5).
    }
    \label{tab:ablation}
        \begin{tabular}{l|c|c|c|c|c|c|c|c|c|c}
            \hline \hline
        \multirow{2}{*}{Methods}  & \multicolumn{2}{c|}{BASE1} & \multicolumn{2}{c|}{BASE2} & \multicolumn{2}{c|}{BASE3} & \multicolumn{2}{c|}{Magrabia} & \multicolumn{2}{c}{Average} \\ \cline{2-11}
                                  & vs. R1      & vs. R2      & vs. R1      & vs. R3      & vs. R1      & vs. R4      & vs. R1        & vs. R5       & vs. R1 & vs. R*  \\ \hline
        \hline
        Ours w/o Div. & 85.69 & 86.02 & 89.00 & 92.63 & 88.44 & 90.30 & 88.42 & 88.28 & 87.89 & 89.31     \\ \hline
        Ours w/o HF & 90.27 & 88.57 & 88.78 & 84.96 & 89.86 & 89.64 & 87.99 & 85.67 & 89.23 & 87.21     \\ \hline
        Ours (HiTTA) & 89.61 & 90.71 & 90.18 & 93.30 & 88.53 & 91.11 & 89.77 & 89.44 & 89.52 & 91.14     \\ \hline
        \hline
        TENT~\cite{wang2020tent}          & 88.14 & 86.30 & 87.93 & 84.98 & 89.57 & 86.47 & 85.46 & 83.26 & 87.78 & 85.25     \\ \hline
        TENT + HF   & 88.81 & 88.92 & 88.13 & 85.70 & 89.64 & 90.01 & 86.24 & 86.64 & 88.21 & 87.82     \\ \hline
        DeYO~\cite{lee2024entropy}          & 89.11 & 88.31 & 88.71 & 84.70 & 89.72 & 88.93 & 87.49 & 85.28 & 88.76 & 86.81     \\ \hline
        DeYO + HF  & 89.41 & 89.39 & 88.36 & 86.38 & 89.63 & 89.87 & 88.18 & 87.47 & 88.90 & 88.28     \\ \hline
        \hline
        1 Lay. $1\times 1$ Conv. & 90.11 & 88.65 & 89.68 & 92.71 & 91.24 & 89.19 & 87.49 & 88.03 & 89.63 & 89.65     \\ \hline
        1 Lay. $3\times 3$ Conv. & 90.28 & 90.42 & 88.78 & 92.86 & 89.72 & 90.83 & 89.10 & 88.86 & 89.47 & 90.74     \\ \hline
        2 Lay. $3\times 3$ Conv. & 89.61 & 90.71 & 90.18 & 93.30 & 88.53 & 91.11 & 89.77 & 89.44 & 89.52 & 91.14     \\ \hline
        \hline
        Entropy as Weight  & 88.92 & 90.28 & 86.56 & 89.68 & 90.01 & 91.04 & 86.64 & 89.10 & 88.03 & 90.03     \\ \hline
        $\mathcal{M}_{div}$ as Weight & 89.61 & 90.71 & 90.18 & 93.30 & 88.53 & 91.11 & 89.77 & 89.44 & 89.52 & 91.14     \\ \hline
        \hline
        \end{tabular}
\end{table}

\section{Conclusion}
In this research, we incorporate the human-in-the-loop technique into TTA. The resultant HiTTA framework, along with a divergence loss, boosts the adaptation of deep learning-based medical image segmentation model to differing data distributions across various medical institutions.
By seamlessly integrating clinician feedback into the TTA process, HiTTA is able to align model predictions more closely to clinical preferences, dramatically enhancing model utility in real clinical settings.
Our extensive experiments on a cross-domain, multi-annotator, and joint OD/OC segmentation dataset underscore the pivotal role that human expertise plays in refining AI model performance. This sets a solid foundation for expanding future work to other medical image segmentation tasks and optimizing human-AI collaboration, thus paving the way for adaptable, efficient AI tools to transform medical diagnostic and treatment processes.

\subsubsection{Acknowledgement:} 
This work was supported in part by the National Natural Science Foundation of China under Grant 62171377, in part by the Key Technologies Research and Development Program under Grant 2022YFC2009903 / 2022YFC2009900, in part by the Natural Science Foundation of Ningbo City, China, under Grant 2021J052, and in part by the Innovation Foundation for Doctor Dissertation of Northwestern Polytechnical University under Grant CX2023016.

\bibliographystyle{splncs04}
\bibliography{reference}

\end{document}